\documentclass[10pt,twocolumn,letterpaper]{article}

\usepackage[pagenumbers]{cvpr}  

\usepackage{capt-of}
\usepackage{xspace}

\usepackage{stfloats}
\usepackage{graphicx}

\usepackage{xcolor}
\usepackage{enumitem}
\usepackage[accsupp]{axessibility}
\usepackage{amsmath,amsthm,amssymb,amsfonts,dsfont,pifont,bm,bbm,mathrsfs,mathtools,nicefrac,extarrows,relsize}
\usepackage{algorithm,algpseudocode,listings}
\usepackage{booktabs,multirow,adjustbox,diagbox,threeparttable,tabularray,setspace}

\definecolor{cvprblue}{rgb}{0.21,0.49,0.74}
\usepackage[pagebackref,breaklinks,colorlinks,citecolor=cvprblue,bookmarks=false]{hyperref}
\usepackage{wrapfig}
\usepackage[capitalize]{cleveref}  

\crefname{section}{Sec.}{Secs.}
\Crefname{section}{Section}{Sections}
\crefname{appendix}{App.}{Apps.}
\Crefname{appendix}{Appendix}{Appendices}
\crefname{table}{Tab.}{Tabs.}
\Crefname{table}{Table}{Tables}
\crefname{figure}{Fig.}{Figs.}
\Crefname{figure}{Figure}{Figures}
\crefname{equation}{Eq.}{Eqs.}
\Crefname{equation}{Equation}{Equations}
\crefname{theorem}{Thm.}{Thms.}
\Crefname{theorem}{Theorem}{Theorems}
\crefname{lemma}{Lem.}{Lems.}
\Crefname{lemma}{Lemma}{Lemmas}
\crefname{remark}{Rem.}{Rems.}
\Crefname{remark}{Remark}{Remarks}
\crefname{corollary}{Cor.}{Cors.}
\Crefname{corollary}{Corollary}{Corollaries}
\crefname{algorithm}{Alg.}{Algs.}
\Crefname{algorithm}{Algorithm}{Algorithms}
\definecolor{cellred}{RGB}{213, 123, 101}
\definecolor{cellgreen}{RGB}{0, 205, 0}
\definecolor{cellblue}{RGB}{54, 125, 189}
\definecolor{codegreen}{rgb}{0,0.6,0}
\definecolor{codegray}{rgb}{0.5,0.5,0.5}
\definecolor{codepurple}{rgb}{0.58,0,0.82}
\definecolor{backcolour}{rgb}{1.0,1.0,1.0}
\lstdefinestyle{mystyle}{
    backgroundcolor=\color{backcolour},
    commentstyle=\color{codegreen},
    keywordstyle=\color{magenta},
    numberstyle=\tiny\color{codegray},
    stringstyle=\color{codepurple},
    basicstyle=\ttfamily\scriptsize,
    breakatwhitespace=false,
    breaklines=true,
    captionpos=b,
    keepspaces=true,
    numbers=left,
    numbersep=5pt,
    showspaces=false,
    showstringspaces=false,
    showtabs=false,
    tabsize=2
}
\newcommand{\tocite}[1]{{\color{red} [TO CITE]}}
\newcommand{\methodname}{LeviTor}
\newcommand{\method}{\texttt{\methodname}\xspace}

\newcommand{\xmark}{\ding{55}}

\title{\methodname: 3D Trajectory Oriented Image-to-Video Synthesis}

\author{Hanlin Wang$^{1,2}$ \quad Hao Ouyang$^{2}$ \quad Qiuyu Wang$^{2}$ \quad Wen Wang$^{3,2}$, \\  
Ka Leong Cheng$^{4,2}$ \quad Qifeng Chen$^{4}$ \quad Yujun Shen$^{2}$ \quad Limin Wang$^{1,5\dagger}$\\[0.3cm] $^1$State Key Laboratory for Novel Software Technology, Nanjing University \quad $^2$Ant Group \\ $^3$Zhejiang University \quad $^4$Hong Kong University of Science and Technology \\ $^5$Shanghai Artificial Intelligence Laboratory 
}

\begin{document}
    
\twocolumn[{
\renewcommand\twocolumn[1][]{#1}
\maketitle
\vspace{-20pt}
\begin{center}
    \includegraphics[width=\linewidth]{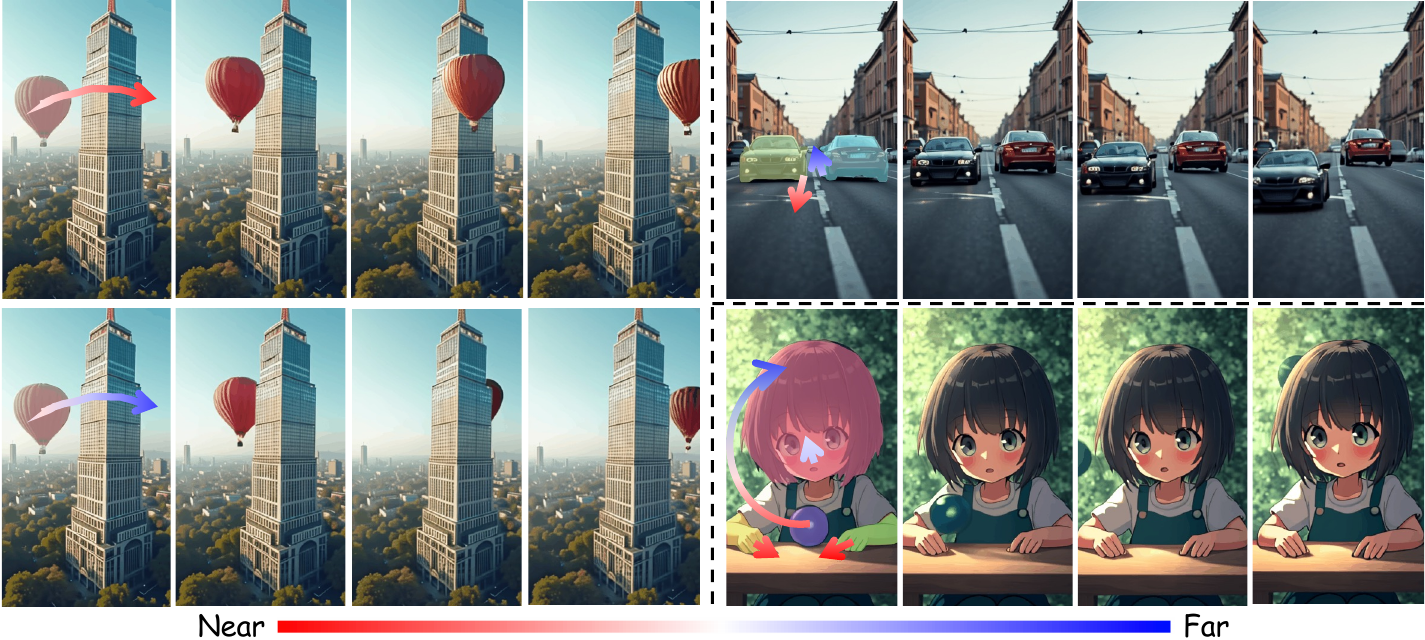}
    \captionsetup{type=figure}
    \vspace{-0.5cm}
    \caption{%
        \method is capable of generating videos with controlled occlusion, better depth changes, and complex 3D orbiting movement based on user inputs. Given an initial frame, users can easily draw 3D trajectory using our inference pipeline to represent their desired movements for designated area. We highly recommend viewing the supplementary materials for detailed video demonstrations.
    }
    \label{fig:teaser}
\end{center}
\vspace{10pt}
}]
\let\thefootnote\relax\footnotetext{\noindent$^\dagger$Corresponding author.}


\begin{abstract}

The intuitive nature of drag-based interaction has led to its growing adoption for controlling object trajectories in image-to-video synthesis.
Still, existing methods that perform dragging in the 2D space usually face ambiguity when handling out-of-plane movements.
In this work, we augment the interaction with a new dimension, \textit{i.e.}, the depth dimension, such that users are allowed to assign a relative depth for each point on the trajectory.
That way, our new interaction paradigm not only inherits the convenience from 2D dragging, but facilitates trajectory control in the 3D space, broadening the scope of creativity.
We propose a pioneering method for 3D trajectory control in image-to-video synthesis by abstracting object masks into a few cluster points. 
These points, accompanied by the depth information and the instance information, are finally fed into a video diffusion model as the control signal.
Extensive experiments validate the effectiveness of our approach, dubbed \method, in precisely manipulating the object movements when producing photo-realistic videos from static images.
Our code is available at:
\textcolor{magenta}{\href{https://github.com/ant-research/LeviTor}{https://github.com/ant-research/LeviTor}}.

\end{abstract}

\section{Introduction}\label{sec:intro}

Controlling object trajectories in video generation~\cite{Motionctrl, DragNUWA, DragAnything, TrackGo} is a fundamental task with wide-ranging applications in computer graphics, virtual reality, and interactive media. Precise trajectory control allows for generation of dynamic scenes where objects move in desired paths, enabling creators to create realistic and compelling visual content. Such control is crucial for tasks like animating characters in a virtual environment, simulating physical phenomena, and developing advanced visual effects that require objects to interact seamlessly within a scene.

Despite its importance, controlling object trajectories in video synthesis presents significant challenges. Traditional methods~\cite{DragNUWA, DragAnything, TrackGo} often rely on 2D trajectory inputs drawn directly on images. While these approaches allow for motion representation to some extent, they inherently suffer from ambiguity and complexities associated with interpreting 2D motions in 3D space. Consider the example of animating a hot air balloon flowing over a building as illustrated in \cref{fig:teaser}. A 2D trajectory drawn on the image cannot distinguish whether the balloon should pass in front of or behind the building. This ambiguity arises because a single 2D path can correspond to multiple 3D trajectories due to the lack of 3D information, making it insufficient for precise control over object movements in a 3D space. However, extracting accurate 3D trajectories poses additional difficulties, especially in scenes with occlusions or complex interactions between objects.  For users, inputting valid 3D trajectories is also non-trivial. It often demands specialized knowledge and tools to define object paths accurately within a 3D space, which can be a barrier for artists and non-expert users aiming to create video content.

To address these challenges, we propose \method, a novel model that fine-tunes pre-trained video generation models to incorporate an efficient and effective 3D trajectory control mechanism. 
Our approach introduces an innovative representation of control signal by combining depth information with K-means clustered points of object masks in video. Such control signal can clearly indicate the occlusion and depth changes between objects through the aggregation or separation of clustered points and their depth.
This fusion also captures essential 3D attributes of objects' trajectory without the need for explicit 3D trajectory estimation, thus simplifying the modeling of complex object motions and interactions. For training, we utilize the recently released high-quality Video Object Segmentation (VOS) dataset from SAM2~\cite{SAM2}, which provides rich annotations conducive to our method. By integrating depth cues with clustered points, our representation effectively encodes the object's spatial movements and depth variations over time. This method not only enhances the model's ability to interpret and generate accurate 3D motions but also mitigates issues related to occlusions and depth ambiguities.

We also design a user-friendly inference pipeline that lowers the barrier for users to input 3D trajectories. Users can simply draw trajectories on 2D images and adjust point depths interactively, which the system then interprets as 3D paths for object movements. This approach streamlines the process, making it accessible to users without extensive technical expertise in 3D modeling or animation. 

Our method demonstrates superior performance both quantitatively and qualitatively compared to existing approaches. We achieve accurate 3D trajectory control in image-to-video synthesis task where previous baselines fail.
In summary, our contributions are as follows: We introduce \method, a novel method for controlling 3D object trajectories in video synthesis by combining depth information with K-means clustered points without the need for explicit 3D trajectory tracking. We leverage the high-quality SA-V dataset for training, effectively capturing complex object motions and interactions in diverse scenes. We develop a user-friendly inference pipeline that simplifies the input of 3D trajectories, making it accessible to a broader range of users. To the best of our knowledge, this work is the first to introduce 3D object trajectory control in image-to-video synthesis, paving the way for more advanced and accessible video generation techniques.

\section{Related Work}\label{sec:related}

\subsection{Video Diffusion Models}

Diffusion models~\cite{first_diffusion, ddpm, ddim} have demonstrated unprecedented power in video generation. Video Diffusion Models (VDMs)~\cite{vdm} are broadly categorized into Text-to-Video (T2V) and Image-to-Video (I2V) frameworks, aiming to generate video samples from text prompts or image prompts. T2V generation~\cite{align_your_latents, videocrafter_1, dreamix, make_a_video, SVD, phenaki, tune_a_video, cogvideo, cogvideox, khachatryan2023, ge2023, animatediff} has been extensively studied in recent years, introducing text descriptions to control the content of video generation semantically. Previous works~\cite{align_your_latents, lvdm, modelscope, lavie, magicvideo, animatediff} incorporate temporal layers into large pretrained text-to-image (T2I) diffusion models~\cite{rombach2022}. Subsequent studies~\cite{align_your_latents, SVD, cogvideox, opensora, opensora_plan} have expanded T2V capabilities by utilizing large text-video pairs, achieving improved results. Building upon T2V, I2V synthesis~\cite{i2vgen_xl, SVD, cogvideox, opensora, videocrafter2, dynamicrafter, liu2024dynamic} has also been widely explored. Given a still image, I2V aims to animate it into a video clip that retains all visual content from the image and exhibits naturally suggested dynamics. Many recent works, such as SVD~\cite{SVD}, VideoCrafter2~\cite{videocrafter2} and CogVideoX~\cite{cogvideox} support both T2V and I2V simultaneously.

Despite producing high-quality videos, these models rely on text or image prompts, limiting fine-grained control and potentially leading to actions misaligned with user intentions. For precise control, some works~\cite{hu2024animate,huang2024lvcd,wang2024framer,zhu2024zero,xu2024magicanimate,ma2024follow,qin2023dancing,zhang2024mimicmotion,xing2023make,zhu2024champ,jeong2023tpos,zhang2025motion} employ multimodal video sequences as conditions, such as pose~\cite{zhu2024zero,xu2024magicanimate,hu2024animate,ma2024follow,qin2023dancing,zhang2024mimicmotion}, depth~\cite{xing2023make,he2023animate,zhu2024champ}, or sound~\cite{jeong2023tpos,lee2023aadiff,liu2023generative,tian2024emo}, treating video generation as a video translation task. Although these models achieve precise control, they require per-frame dense control signals, which makes them cumbersome and not user-friendly in real-world applications. Therefore, simpler yet precise control mechanisms are needed. Trajectory-based control offers an effective method for manipulating video generation, combining simplicity with precision.

\subsection{Trajectory Control in Video Generation}

Controllable editing has gained advancements in the field of image editing due to its precise control information~\cite{EpsteinJPEH23,MuGZSVWP24,YenphraphaiPLPX24,BhatMW24}. For video synthesis, trajectory-controlled generation has recently gained popularity due to its ability to achieve precise motion control. Early works~\cite{hao2018controllable, ardino2021click, blattmann2021understanding, ipoke} employed recurrent neural networks or optical flow to guide motion. Methods like TrailBlazer~\cite{trailblazer} utilize bounding boxes to direct subject motion in video generation. MotionCtrl~\cite{Motionctrl} encodes trajectory coordinates into dense vector maps, and DragNUWA~\cite{DragNUWA} transforms sparse strokes into dense flow spaces; both use these representations as guidance signals. Tora~\cite{Tora} employs a motion variational autoencoder~\cite{vae} to embed trajectory vectors into the latent space, preserving motion information across frames.

Although these methods facilitate trajectory control, they often lack semantic understanding of entities, making control over video generation less refined. To address this issue, DragAnything~\cite{DragAnything} combines entity representation extraction with a 2D Gaussian representation to achieve entity-level controllable video generation. TrackGo~\cite{TrackGo} uses user-provided free-form masks and arrows to define target regions and movement trajectories, serving as precise blueprints for video generation. However, all these methods consider 2D trajectories in image space, leading to ambiguities in real 3D environments. The recent 3D-TrajMaster~\cite{3dtrajmaster} manipulates multi-entity 3D motions with user-desired 6DoF pose sequences of entities for video generation. In this paper, we introduce an innovative control signal representation that combines depth information with K-means clustered points from object masks in video, achieving accurate entity-level and 3D trajectory control.

\section{Method}\label{sec:method}
\label{sec::method}

\begin{figure}[t]
  \centering
  \includegraphics[width=\linewidth]{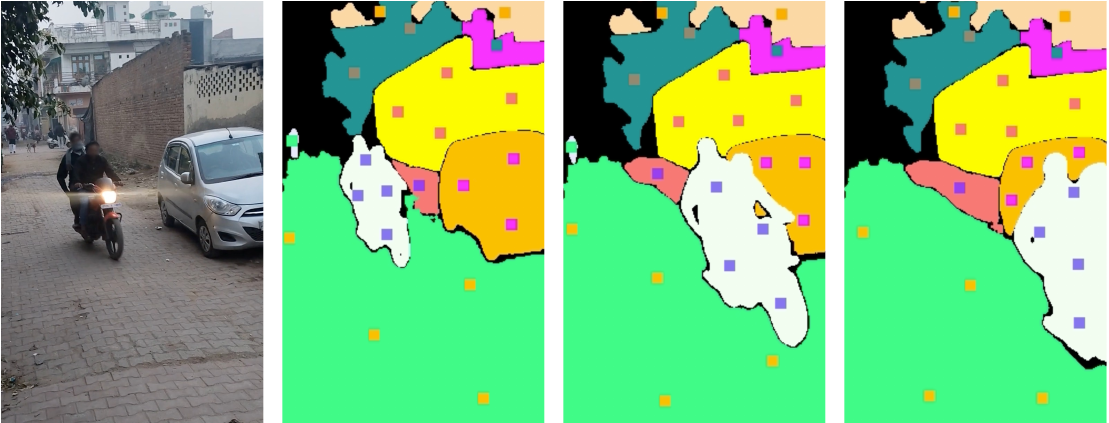} 
   \caption{An example of object movement and occlusion represented by K-means clustered points.}
   \label{fig:motivation}
   \vspace{-15pt}
\end{figure}

\subsection{Problem Formulation}

To learn realistic object motion, the training dataset should contain high-quality videos with accurate object motions. However, existing datasets that provide 3D motion trajectories are either limited in size or consist solely of synthetic data. The Video Object Segmentation (VOS) datasets~\cite{davis, SAM2}, particularly with the recent release of SAM2~\cite{SAM2}, offer high-quality videos with precise object mask annotations, making it an appropriate choice for our purposes. Nevertheless, two primary challenges remain:

\begin{enumerate}
    \item The dataset lacks explicit 3D trajectory information, which is essential for training a model to understand and synthesize 3D motions. Therefore, we need to implicitly express the 3D motion information contained in the data.
    \item The provided mask annotations are too detailed for practical user input, as users cannot be expected to supply such fine-grained masks or dense 3D trajectories for control. Thus it is necessary to design a representation of 3D trajectories that is easy for users to input.
\end{enumerate}

To address these issues, we propose using K-means points extracted from the object masks along with their depth information as control signals. Specifically, we apply K-means clustering to the pixels of the mask to obtain a set of representative control points:
\begin{equation}
\left\{ (x_t^i, y_t^i) \right\}_{i=1}^N = \text{K-means}(M_t, N),
\end{equation}
where $M_t$ denotes all object masks at frame $t$, $N$ is the number of clusters (control points), and $(x_t^i, y_t^i)$ is the 2D coordinate of control point $i$ at frame $t$.
These control points not only simplify user input but also encapsulate implicit 3D information. As illustrated in Figure~\ref{fig:motivation}, the spatial distribution and density of the K-means points reflect changes in the object's depth and motion. For example, as a motorbike moves closer to the camera, the points spread out due to perspective scaling, indicating depth changes. Similarly, during occlusions, the distribution of points on the car shifts, capturing the occlusion dynamics.
Then we employ a depth estimation network, DepthAnythingV2~\cite{depth_anything_v2}, to predict relative depth maps $\{ D_t \}_{t=1}^L$ for frames in the dataset, where $L$ is the video length. In this way, we avoid the need of absolutely accurate depth information, making it easier for users to interact. We sample the depth at each control point:
\begin{equation}
d_t^i = D_t(x_t^i, y_t^i),
\end{equation}
where $d_t^i$ is the depth value at control point $i$ in frame $t$. This process enriches the control points with depth information, effectively providing approximate 3D coordinates without requiring explicit 3D annotations. By combining the 2D coordinates and the estimated depth values, we construct the control trajectories:
\begin{equation}
    \mathcal{T} = \left\{ \left\{ \left( x_t^i, y_t^i, d_t^i \right) \right\}_{t=1}^L \right\}_{i=1}^N,
\end{equation}
This representation allows users to efficiently specify 3D trajectories by simply selecting points on a 2D image and adjusting depth values as needed. We thus design our training and inference pipeline as in \cref{method:training} and \cref{method:inference}.

\begin{figure}[t]
  \centering
  \includegraphics[width=\linewidth]{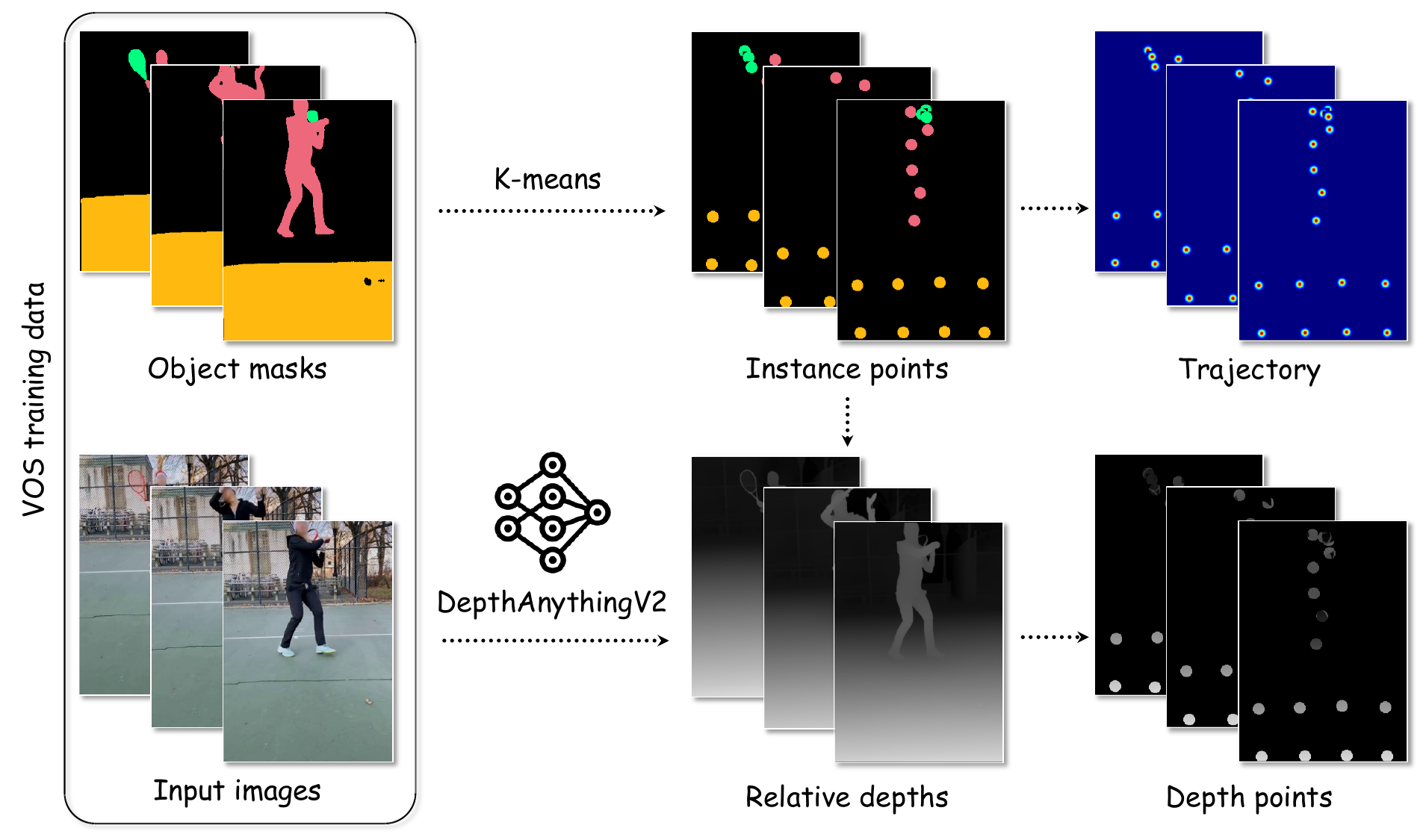} 
   \caption{Control signal generation process of \method.}
   \label{fig:control}
   \vspace{-15pt}
\end{figure}
\begin{figure*}[t]
  \centering
  \includegraphics[width=\linewidth]{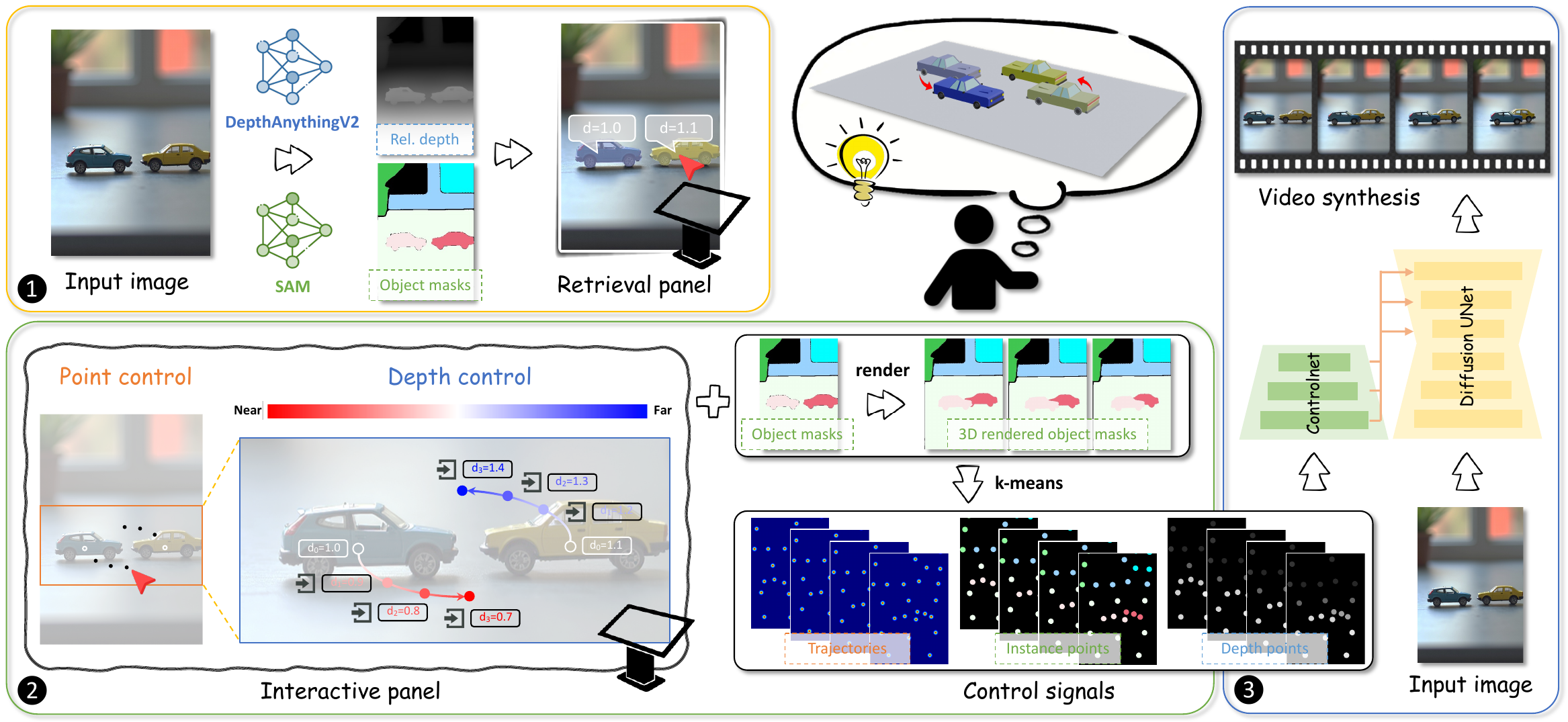} 
   \caption{Inference pipeline of \method, which consists of user retrieval panel, interactive panel, 3D rendered object masks generation and video synthesis. Users can easily draw 3D trajectories through our retrieval panel and interactive panel, and our system later use these inputs to generate user desired videos.}
   \label{fig:infer_pipeline}
   \vspace{-10pt}
\end{figure*}

\subsection{Training Pipeline}
\label{method:training}

Given a VOS format video $V \in \mathbb{R}^{L \times H \times W \times 3}$, it provides the ground truth masks of multiple objects in the video, represented as $\{\{M_i^j\}_{i=1}^X\}_{j=1}^L$, where $X$ denotes the number of object masks in each frame. For each mask $M_i^j$, we conduct K-means algorithm to obtain $k$ center points as control signal. Specifically, we first calculate the area ratio of $M_i^j$ to the entire image and multiply a hyper-parameter $\alpha$ to determine the approximate number of cluster points:

\vspace{-5pt}
\begin{equation}
k = (\frac{S_{M_i^j}}{H*W}) * \alpha
\end{equation}

Then we assess whether there is a significant change of $S_{M_i^j}$, which indicates 3D related situations such as the object being occluded, moving out of the frame, or changing distance from the lens. To achieve this, we go through all video frames and calculate the ratio of the maximum to minimum area of the $i_{th}$ object. If the ratio exceeds 10, we ensure that the value of $k$ is not less than 3 in order to better represent the changes of this object along the temporal dimension:

\vspace{-10pt}
\begin{equation}
k = 
\begin{cases}
{\rm max}(k, 3), & {\rm if} \;\; \frac{{\rm max}(\{S_{M_i^j}\}_{j=1}^L)}{{\rm min}(\{S_{M_i^j}\}_{j=1}^L)} > 10, \\
k, & {\rm otherwise},
\end{cases}
\end{equation}

We later ensure $k \le 8$ to avoid the issue of having too many control points. We perform K-means clustering with the calculated $k$ value on $M_i^j$ and use the resulting cluster centers as control points. After extracting key points for all objects in each frame, we obtain the 2D coordinate information of all control points and instance information that show which object the point belongs to.

We then use DepthAnythingV2~\cite{depth_anything_v2} to estimate the relative depth of each frame. Thus we can assign depth value to the corresponding 2D coordinate trajectories to get 3D trajectories. Finally, we represent the 2D trajectories with Gaussian heatmap and concatenate the trajectories, instance points, and depth points to serve as control signal, which is injected into the Stable Video Diffusion (SVD)~\cite{SVD} using ControlNet~\cite{controlnet} to generate a video that aligns with the 3D trajectory. Our control signal generation process is shown in \cref{fig:control}.

Our training process can be represented as:
\vspace{-5pt}
\begin{equation}
\mathcal{L}=\mathbb{E}_{z_t, z^0, t, \epsilon \sim \mathcal{N}(0, \mathbf{I})}\left[\left\|\epsilon-\epsilon^{c}_{\theta}\left(z_t; t, z^0, c_{traj}\right)\right\|^2\right],
\label{eq:conditonal_denoising}
\end{equation}
where $z^0$ denotes VAE-encoded latent feature of the first frame, $c_{traj}$ means the control signal and $\epsilon^{c}_{\theta}$ is the combination of the denoising U-Net and the ControlNet branch.

\subsection{Inference Pipeline}
\label{method:inference}

\begin{figure}[t]
  \centering
  \includegraphics[width=\linewidth]{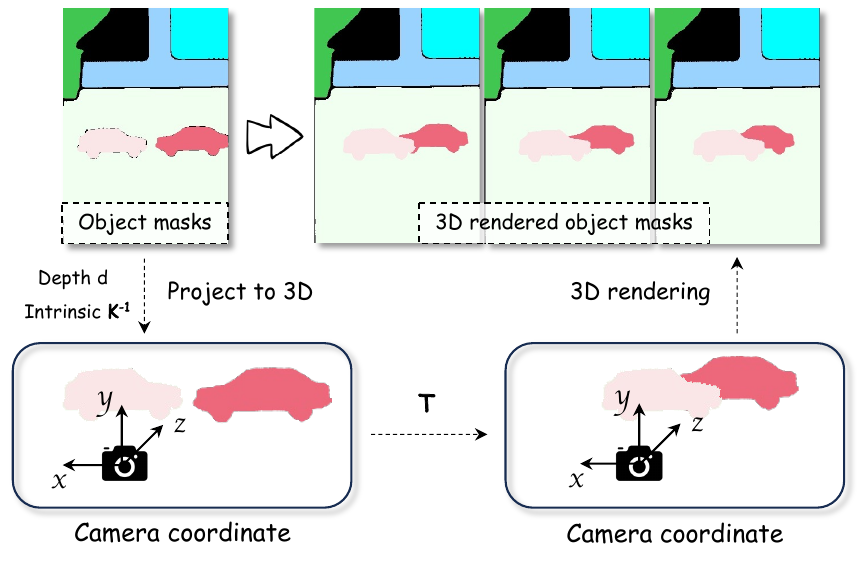} 
   \caption{3D rendered object masks generation pipeline.}
   \label{fig:3Drender}
   \vspace{-15pt}
\end{figure}

We have designed a user-friendly interactive system for inference and the overview is provided in \cref{fig:infer_pipeline}. Take an image as input, the system first automatically extracts depth information and object masks from the image using DepthAnythingV2 and SAM. Then users can utilize the retrieval panel to select the masks of objects to be moved by simply clicking on the image. They can also get relative depth values of clicked points automatically. After that, the user can use the interactive panel to click on more points to form the object trajectory. At the same time, the user can refer to the relative depth values of previously obtained click positions to input depth information of points within the trajectory according to their needs, thereby providing the corresponding 3D trajectories. 

With the sparse 3D trajectories and selected masks provided by user as input, we need to convert it into corresponding multi-points control information. This is because requiring users to input multiple point trajectories that comply with physical laws to represent correct occlusions and depth changes is hard. Generally, they only input a single trajectory to indicate the movement of an object. Thus we need this conversion to represent the 3D movements of objects through the clustering or dispersion of control points.  We achieve this by generating 3D rendered object masks then selecting control points with K-means, as illustrated in \cref{fig:3Drender}. Specifically, we first combine the 2D coordinates of pixels in the starting image with their depth values to obtain 3D spatial points, represented as $\{P_i\}_{i=1}^n = \{x_i, y_i, d_i\}_{i=1}^n$, where $n$ means the number of pixels in selected masks. Then we transform these points into the camera coordinate system. We assume that all camera intrinsic parameters are all the same and the camera to be still, so the rotation matrix is an identity matrix. The first step of transformation is converting 2D pixel points with their depth value into the camera coordinate system and moving the points belonging to user selected masks in this transformed 3D space:

\vspace{-10pt}
\begin{equation}
    \begin{split}
        [X_i, Y_i, Z_i]^T &= \mathbf{K}^{-1} \cdot [x_i, y_i, 1]^T \cdot d_i, \\
        [X'_i, Y'_i, Z'_i]^T &= [X_i, Y_i, Z_i]^T + \mathbf{T},
    \end{split}
\end{equation}
here $\mathbf{K}$ denotes the perspective projection matrix of camera and $\mathbf{T}$ is the moving vectors assigned by users. After that, we render these points back to 2D images: 

\vspace{-5pt}
\begin{equation}
    [x_i, y_i]^T = f\left ( [X'_i, Y'_i, Z'_i]^T, {ID}_i) \right ),
\end{equation}
$f$ is a rendering function which we implement with renderer function in PyTorch3D~\cite{pytorch3d} and $ID_i$ is the instance that the $i_{th}$ point belongs to. All the points are assigned the corresponding instance information, so rendering them back results in images with masks of different objects.

In this way, we represent the movements, occlusion, and size changes due to forward and backward movements of objects only with the sparse trajectories input by the user. At the same time, the changes in 2D masks rendered from 3D space also fully comply with the laws of physics.

By mapping points to 3D space and then rendering them back to 2D mask images, we convert sparse user controls into dense mask representations. These masks can accurately reflect the movement and occlusion of objects. Next, we compute cluster centers using K-means based on the masks obtained from rendering. By combining these with user-specified depth changes, we derive an appropriate number of control trajectories to generate the final video using our \method. Further selecting control points with K-means is necessary because the movement process in 3D space cannot represent non-rigid transformations. If we directly use a dense mask for control, it will only result in a straightforward translation of the object, as demonstrated in \cref{fig:ablation2}. By converting the mask into a moderate number of trajectory control signals, the generative model can capture the motion variation of the object while also adding some details of non-rigid movements.

\section{Experiments}
\label{sec:exp}
\begin{figure*}[htbp]
  \centering
  \includegraphics[width=\linewidth]{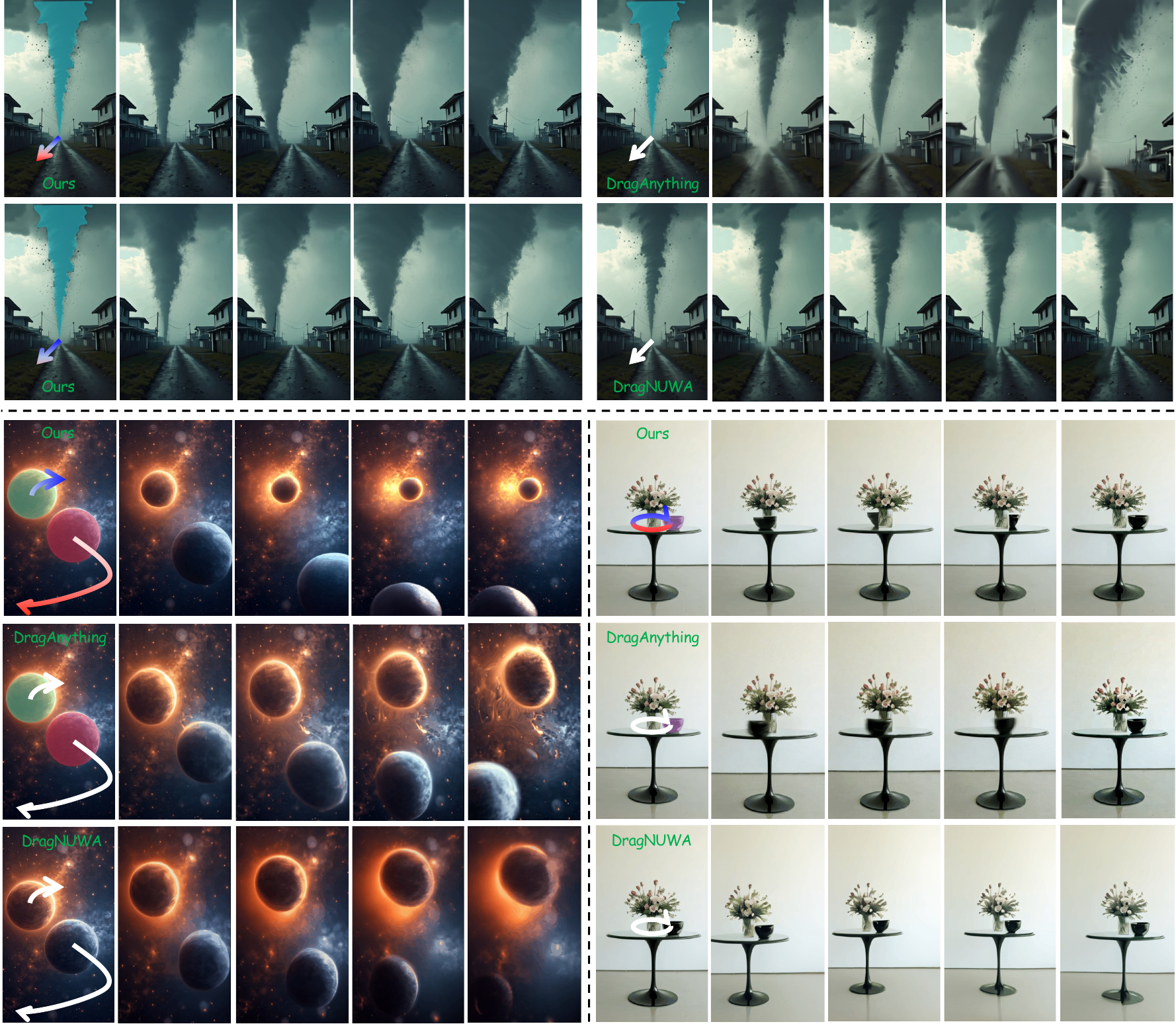} 
   \caption{Qualitative comparison with DragAnything~\cite{DragAnything} and DragNUWA~\cite{DragNUWA}.  \method and DragAnything both support moving user-selected mask areas, whereas DragNUWA directly encodes trajectories as control signals and does not support user selection of operation areas. The top two rows show evaluation on control of mutual occlusion between objects. The left bottom images show comparison of forward and backward object movements control. The right bottom images show a case of complex motion implementation.}
   \label{fig:comp}
   \vspace{-10pt}
\end{figure*}

\subsection{Experiment Settings}

\noindent \textbf{Implementation details.}
We use SVD~\cite{SVD} as our base model. During training, we sample 16 consecutive frames from videos at a spatial resolution of $288$×$512$. Specifically, we center-crop the video to an aspect ratio of $288/512$, then resize the video frames to the resolution of $288$×$512$.
Our \method is trained for 200K iterations using the AdamW optimizer with a learning rate of 1e-5. All training is conducted on 16 NVIDIA A100 GPUs with a total batch size equal to 16.

\noindent \textbf{Datasets.}
For training, we utilize the high-quality Video Object Segmentation (VOS) dataset Segment Anything Video (SA-V)~\cite{SAM2}, which consists of 51K diverse videos and 643K high-quality spatio-temporal segmentation masks. We conduct an evaluation on the DAVIS~\cite{davis} dataset and split videos into clips with 16 frames for testing. Inspired by DragAnything~\cite{DragAnything}, we apply K-means to the mask of each object in the start frame to select K points in each mask area as control points. Then, we employ Co-Tracker~\cite{co-tracker} to track these control points to generate corresponding point trajectories as the ground truth.

\noindent \textbf{Metrics.}
Following ~\cite{DragAnything, Motionctrl}, we adopt Frechet Video Distance~\cite{FVD} (FVD) to measure video quality and assess image quality using Frechet Inception Distance~\cite{FID} (FID). For motion controllability evaluation,
we leverage ObjMC~\cite{Motionctrl}, which computes the Euclidean distance between the generated and pre-defined trajectories. Trajectories of generated videos are extracted using Co-Tracker.

\subsection{Comparison with Other Approaches}

We compare our methods with DragNUWA~\cite{DragNUWA} and DragAnything~\cite{DragAnything}, which enable motion control on given images and have publicly available code. We conduct both qualitative and quantitative comparisons.

\noindent \textbf{Qualitative comparison.}
For qualitative analysis, we focus on verifying the crucial role of introducing 3D trajectories into video generation, which includes the following three aspects: 1) The control of mutual occlusion between objects; 2) Better control for forward and backward object movements in relation to the lens; 3) The implementation of complex motions (such as orbiting). 

Qualitative comparison results are shown in \cref{fig:comp}, where we input the same 2D control trajectory to all models. The top two rows of images show the verification results of occlusion control. In this case, we provide our \method with different depth variations: the depth in the first row changes from far to near, while the depth in the second row only moves closer without being closer to the camera than the buildings on street side. The generated results perfectly meet our requirements, with the tornadoes progressing from far to near and gradually getting larger. Meanwhile, tornado in the first row sweeps across the front of the building, while in the second row it just passes behind the building. In contrast, the other two methods can only control the generation through 2D trajectories. It can be observed that DragAnything misinterprets the movement of the tornado as a forward movement of the camera, resulting in a blurry output. On the other hand, DragNUWA correctly understands that the tornado needs to be moved. However, since it lacks consideration of changes in depth, the size of the tornado hardly changes after the movement, which does not comply with perspective projection rules.

Evaluation results on control for forward and backward object movements in relation to the lens are shown as the left-bottom images in \cref{fig:comp}. It is clear that 2D trajectory cannot provide depth information, so DragAnything and DragNUWA can only simulate planets motion that conforms to that trajectory, resulting in blurry videos. In contrast, \method can generate accurate and clear movements of two planets based on user-specified inputs meanwhile conforming to perspective projection rules.

Based on the information input by users, we can derive 3D trajectories to control the movement of objects, which represent users' desired object occlusions and size changes. Furthermore, we can simulate more complex motions, such as object orbiting. The right-bottom images in \cref{fig:comp} shows an example and our model is able to accurately simulate the situation of a black bowl rotating around a vase and correctly handle the occlusion relationships. Instead, DragAnything cannot directly interpret the 2D trajectory to achieve our desired swirling effect. It only generates a video where the bowl moves from right to left and then back. During this movement, the bowl also undergoes distortion and blurring. DragNUWA treats this 2D input as a camera trajectory, resulting in a video that shows a stationary table and bowl filmed from different angles.

\begin{figure}[t]
  \centering
  \includegraphics[width=\linewidth]{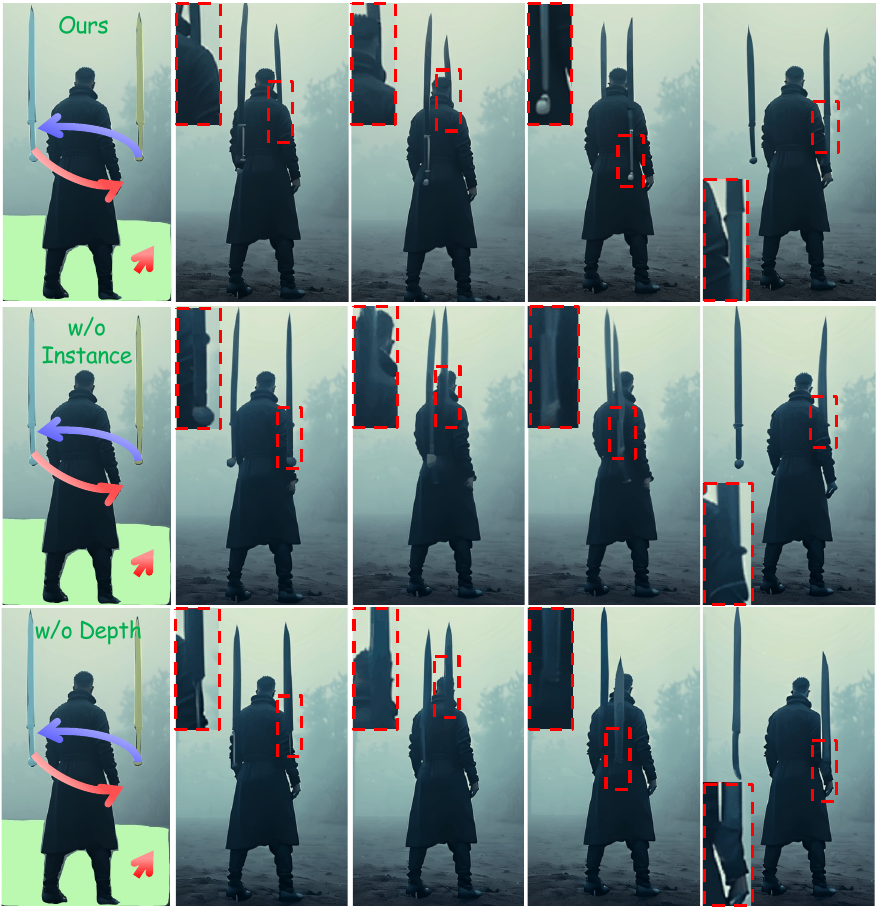} 
   \caption{Ablation on Instance and Depth information. Enlarged details are shown in red boxes. Zoom in for better viewing.}
   \label{fig:ablation1}
   \vspace{-5pt}
\end{figure}

The qualitative comparison results demonstrate that by introducing 3D trajectory control which allows for easy input by users, our \method can better manage the proximity changes of objects. It can also produce video results that cannot be generated with only 2D trajectories, such as controlling object occlusion and executing complex movements like orbiting. Additionally, since our pipeline includes all object masks automatically extracted by SAM~\cite{SAM}, \method ensures that only objects selected by users can be moved. This prevents interpreting object movement as camera movement. And camera movement can be implemented by moving the mask of the selected background (as shown in \cref{fig:ablation1}).

\begin{table}[t]
\centering
\caption{
    Quantitative comparison on DAVIS~\cite{davis}.
    }
\resizebox{\linewidth}{!}{
    \begin{tabular}{clcccc}
    \toprule
    Settings & Methods & FID$\downarrow$ & FVD$\downarrow$ & ObjMC$\downarrow$ \\
    \midrule
    & DragAnything~\cite{DragAnything}    & 36.69 & 327.41 & 42.19 \\
    & DragNUWA1.5~\cite{DragNUWA}  &  44.82 & 330.17 & \textbf{33.03} \\
    \cmidrule{2-5}
    \multirow{-3.5}{*}{\rotatebox[origin=c]{0}{Single-Point}} 
    & \method (Ours) & \textbf{28.79} & \textbf{226.45} & 37.39 \\
    \midrule
    & DragAnything~\cite{DragAnything}    & 36.04 & 324.95 & 38.86 \\
    & DragNUWA 1.5~\cite{DragNUWA}  & 42.34 & 299.96 & \textbf{23.12} \\
    \cmidrule{2-5}
    \multirow{-3.5}{*}{\rotatebox[origin=c]{0}{Multi-Points}} 
    & \method (Ours) & \textbf{25.41} & \textbf{190.44} & 25.97 \\
    \bottomrule
    \hline
    \end{tabular}
}
\label{tab:comparison}
\vspace{-15pt}
\end{table}

\noindent \textbf{Quantitative comparison.}
We evaluate the quantitative results with two input settings: Single-Point and Multi-Points. The setting of Single-Point is consistent with previous work~\cite{DragAnything}, which means that only one point trajectory is selected for each mask as video generation condition.
For Multi-Points setting, we select at most 8 points in each mask and use their trajectories as condition. \cref{tab:comparison} shows the quantitative comparison results of \method with baselines on DAVIS. Using the same SVD as base model, our method achieves a significant advantage in both FID and FVD metrics, thanks to the consideration of 3D trajectory and training on high-quality SA-V dataset. Besides, increasing the number of control trajectories can effectively benefit DragNUWA and \method. This indicates that considering object size changes over time and occlusion is effective. DragAnything is trained using a single trajectory with object mask semantic information in first frame, thus increasing the number of trajectories doesn't match the training and improvement is limited. \method performs worse than DragNUWA on the ObjMC metric, which we attribute to the fact that we do not use tracking methods to obtain complete point trajectories and require the generated video to perfectly match these trajectories.

\subsection{Ablation Studies}
We conduct ablations to study how depth points, instance information and the number of control points for inference affect our synthesis results with the Multi-Points setting.

\noindent \textbf{Depth and instance information.}
\cref{tab:ablations} shows the results of training \method without depth or object instance input, which suggest that both depth and instance information are helpful to our model learning. Compared to depth information, object instance is more important because it represents the objects corresponding to different control points. Without this information, model can easily confuse the control points of different objects, leading to blurred and unrealistic results. Depth information of objects is to some extent implicit in the degree of clustering of points, so its impact is relatively small. We also present a qualitative ablation result in \cref{fig:ablation1}, which suggests that without instance or depth information, the model can easily confuse occlusion relationship between objects, resulting in blurry and unrealistic generation results.

\begin{figure}[t]
  \centering
  \includegraphics[width=\linewidth]{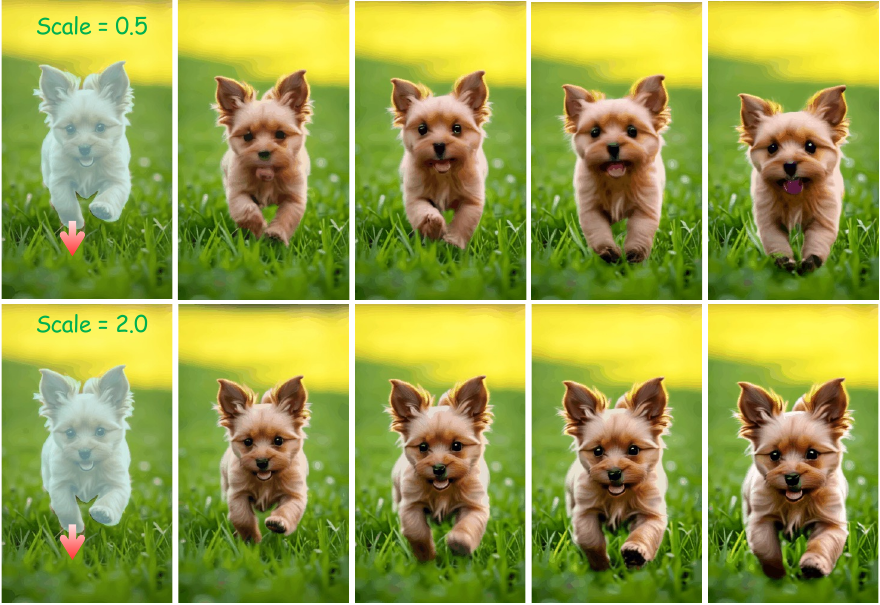} 
   \caption{Ablation on number of inference control points. 'Scale' means the value multiplied by the default number of control points.
   }
   \label{fig:ablation2}
   \vspace{-15pt}
\end{figure}    


\begin{table}[t]
\centering
\caption{
    Ablations on Object Instance and Depth information.
    }
\vspace{-5pt}
\resizebox{0.7\linewidth}{!}{
    \begin{tabular}{cc|cccc}
    \toprule
    Depth & Instance & FID$\downarrow$ & FVD$\downarrow$ & ObjMC$\downarrow$ \\
    \midrule
    \xmark & \xmark & 27.83 & 227.58 & 29.82 \\
    \checkmark & \xmark & 28.04 & 221.29 & 29.13 \\
    \xmark & \checkmark  & 25.45 & 199.44 & \textbf{25.40} \\
    \midrule
    \checkmark & \checkmark  & \textbf{25.41} & \textbf{190.44} & 25.97 \\
    \bottomrule
    \hline
    \end{tabular}
}
\label{tab:ablations}
\vspace{-12pt}
\end{table}

\noindent \textbf{Number of control points for inference.}
During inference, our model can choose different number of control points to strike a balance between motion amplitude and generation quality. \cref{fig:ablation2} illustrates an example, where we multiply the initial number of control points by a scale to evaluate the impact of different numbers of control points on generation results. It can be seen that when there are few control points, the generated result exhibits significant movement amplitude, but the object may experience some deformation or blurring during the motion. 
However, too many control points can get close to the object's mask. Although taking these points as control ensures the reasonableness of the object's shape, it prevents the model from generating the result of its movement. As shown in the last row of \cref{fig:ablation2}, the puppy will translate directly from back to front. Users can therefore adjust the number of control points according to their needs to achieve the desired generation results.
\section{Conclusion}\label{sec:conclusion}
In this paper, we have presented \method, a novel model for implementing 3D object trajectory control in image-to-video synthesis.
Taking depth information combined with K-means clustered points as control signal, our approach captures essential 3D attributes without the need for explicit 3D trajectory estimation. Our user-friendly inference pipeline allows users to input 3D trajectories by simply drawing on 2D images and adjusting point depths, making the synthesis process more accessible. Our model also has certain limitations. First, \method is constrained by the segmentation results of SAM and trajectories provided by the user, and it does not understand physical laws to generate movements of objects without provided trajectory controls. Additionally, since \method was not trained using tracking data, it cannot control the internal pose changes of objects. Finally, the current generation results are limited by the base model SVD. For future work, we aim to extend \method by incorporating more advanced video base models capable of capturing deformable objects and intricate dynamics to better handle non-rigid motions.

\noindent\textbf{Acknowledgements:} This work is supported by the National Key R$\&$D Program of China (No. 2022ZD0160900), the Research Grant Council of the Hong Kong Special Administrative Region under grant number 16203122, Ant Group Research Intern Program, Jiangsu Frontier Technology Research and Development Program (No. BF2024076), and the Collaborative Innovation Center of Novel Software Technology and Industrialization.
\clearpage
\appendix
\renewcommand\thesection{\Alph{section}}
\renewcommand\thefigure{S\arabic{figure}}
\renewcommand\thetable{S\arabic{table}}
\renewcommand\theequation{S\arabic{equation}}
\setcounter{figure}{0}
\setcounter{table}{0}
\setcounter{equation}{0}
\setcounter{page}{1}
\maketitlesupplementary

\section*{Appendix}

\section{Comparison with more methods}
\begin{figure}[h]
\vspace{-8pt}
  \centering
  \includegraphics[width=\linewidth]{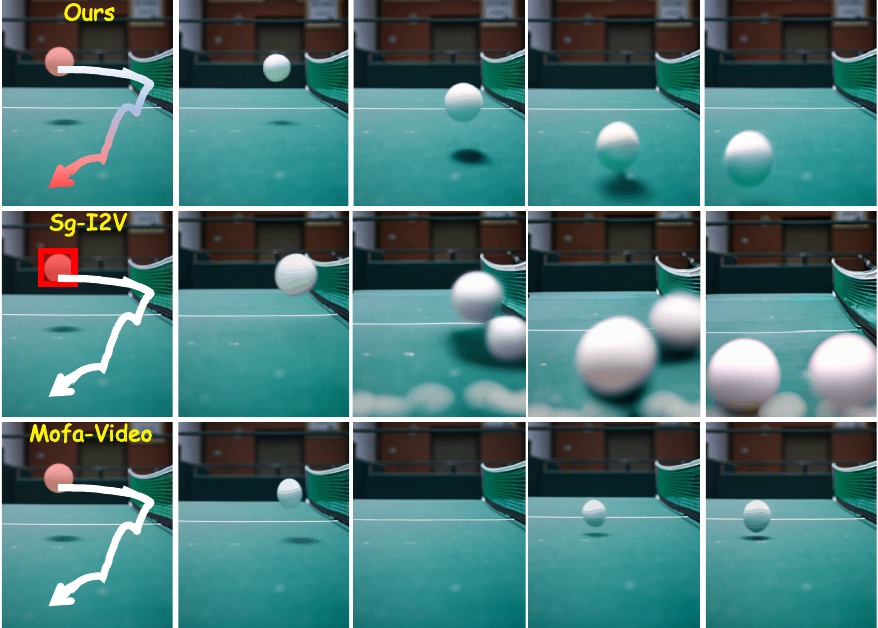}
  \caption{Qualitative comparison with SG-I2V and MOFA-Video.}
  \vspace{-18pt}
  \label{fig:supp0}
\end{figure}

This section compares our \method with more recent methods SG-I2V~\cite{sgi2v} and MOFA-Video~\cite{mofa}. The qualitative comparison in ~\cref{fig:supp0} shows that these methods fail to follow complex trajectories or produce proper depth variation.

\section{More Ablations on the Number of Control Points for Inference}\label{sec:points_number}
\begin{figure}[ht]
    \centering
    \includegraphics[width=\columnwidth]{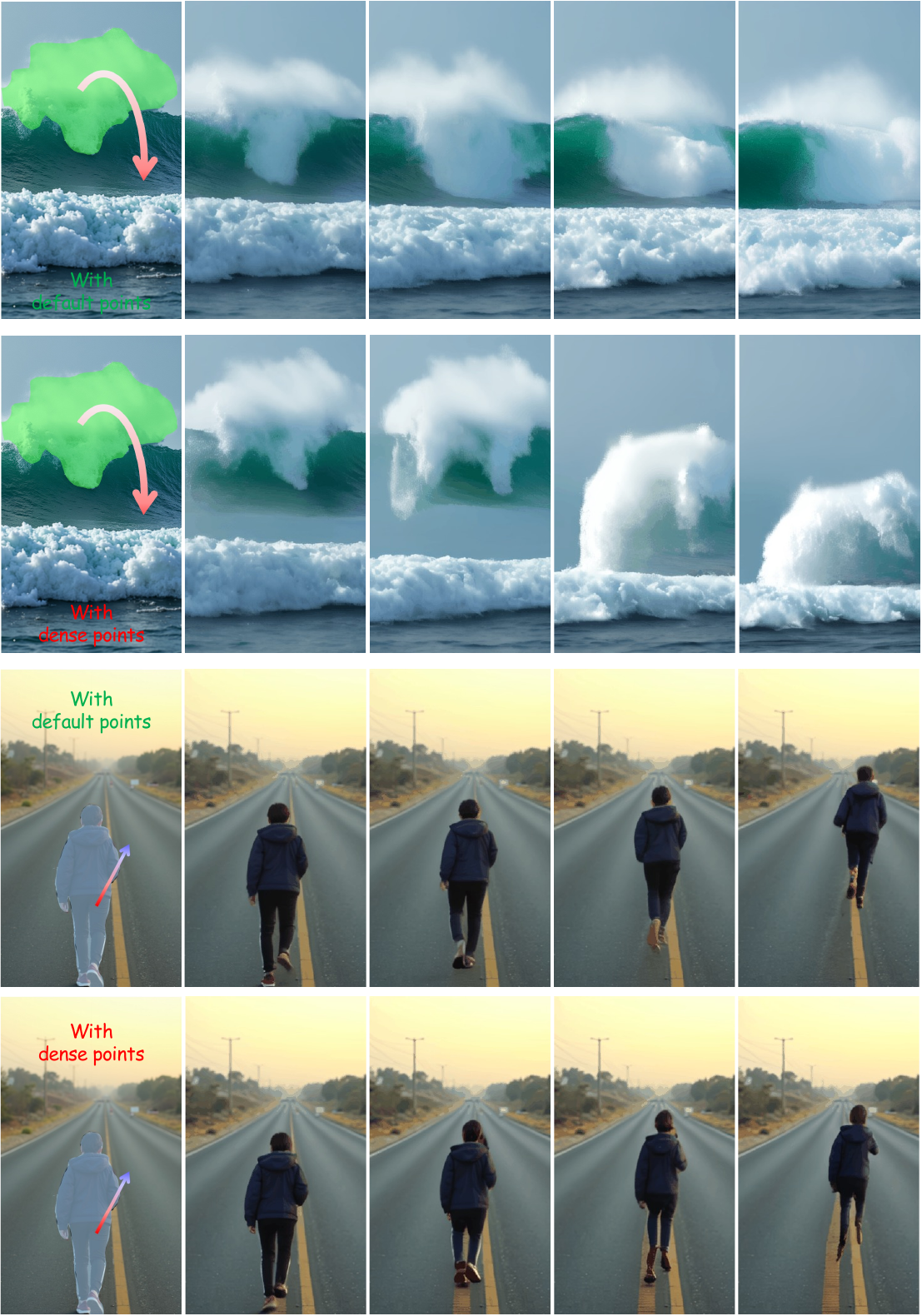}
    \caption{Ablation results on the Number of Control Points for Inference. We highly recommend viewing the visualization results for detailed video demonstrations.}
    \label{fig:supp1}
    \vspace{-5pt}
\end{figure}

In this section, we show more examples of choosing different numbers of control points to generate videos with~\method. We conduct inference with our default number of control points and with more densely packed points, respectively. The results are shown in~\cref{fig:supp1}. It can be seen that with the default number of control points, our~\method can reasonably represent the state of fluid movement and human running. However, since the generation strictly follows the control points, the more control points used, the less space is left for our model to produce some non-rigid movements, resulting in the unreasonable results of waves floating in the air and people gliding on the road. This demonstrates that overly dense control points cannot generate non-rigid motion well. Thus, we implement~\method with multiple clustered points control rather than directly using object masks as the condition. In this way, users can flexibly adjust the number of control points as needed to generate both rigid and non-rigid motions.

\begin{table}[ht]
\centering
\caption{
    Quantitative comparison with Single-point Control on DAVIS~\cite{davis}.
    }
\resizebox{0.9\linewidth}{!}{
    \begin{tabular}{cccc}
    \toprule
    Methods & FID $\downarrow$ & FVD $\downarrow$ & ObjMC $\downarrow$ \\
    \midrule
    Single-Point Control  &  30.91  & 253.73 & 38.21  \\
    Ours  &  \textbf{25.41} & \textbf{190.44} & \textbf{25.97} \\
    \hline
    \end{tabular}
}
\label{tab:supp1}
\vspace{-5pt}
\end{table}

\begin{figure}[t]
    \centering
    \includegraphics[width=\columnwidth]{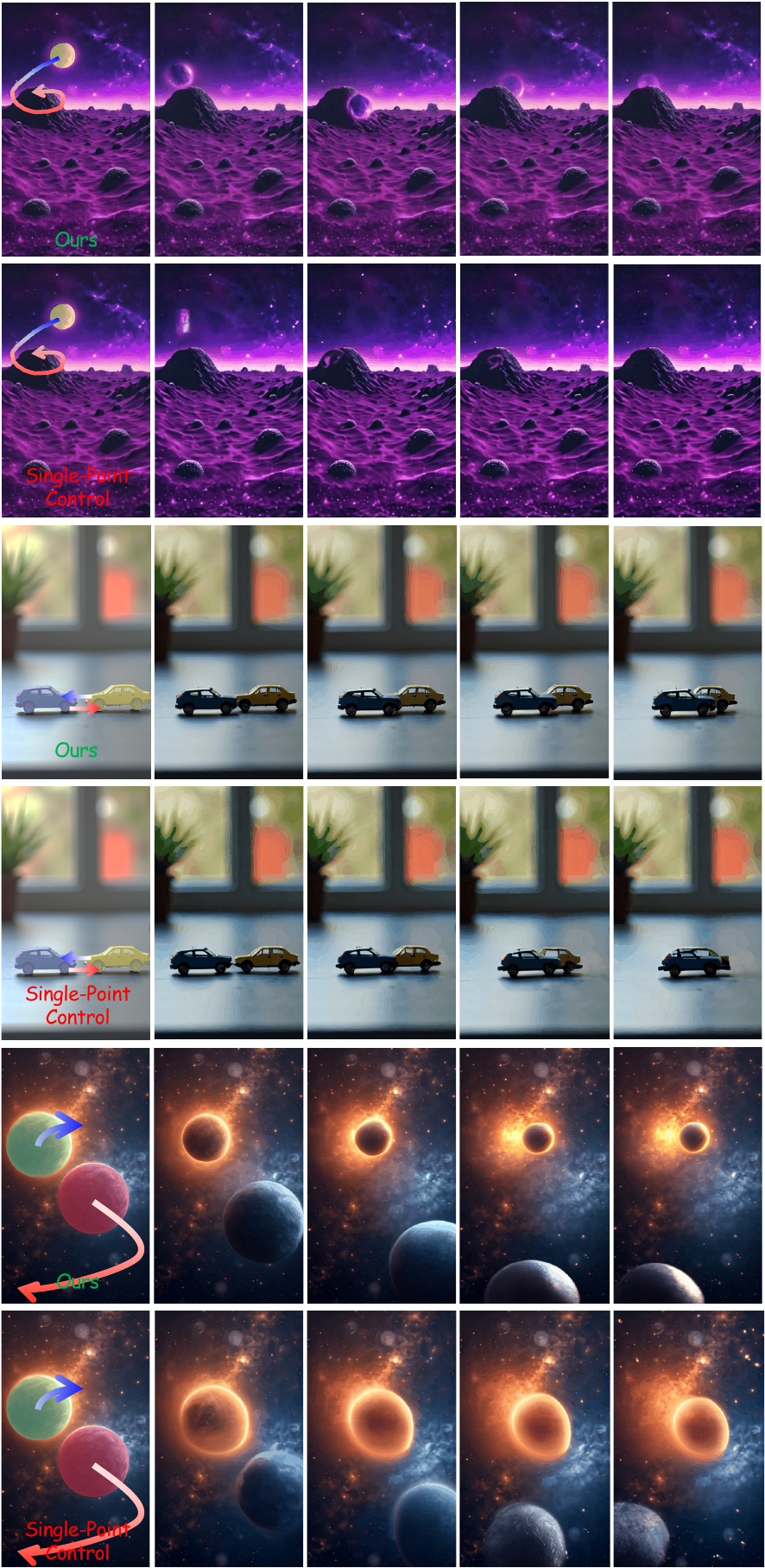}
    \caption{Comparison with Single-point Control model. We highly recommend viewing the visualization results for detailed video demonstrations.}
    \label{fig:supp2}
    \vspace{-10pt}
\end{figure}

\section{Comparison with Single-point Control}\label{sec:single_point}
One of our key motivations is to represent 3D motions by utilizing the clustering and dispersion of multiple points within object masks. Another more intuitive idea is whether we can represent 3D motion using 2D trajectories combined with depth information. That is, representing a 3D trajectory through a single 2D trajectory along with changes of depth values input by users. To validate this idea, we use the center point of each object's mask as a control point and train the model with the value change of that point as the generation condition. We conduct both qualitative and quantitative analysis. Qualitative results in~\cref{fig:supp2} show that such single-point control can not represent 3D motions well. The first two examples test the representation of occlusion. It can be observed that a single point with depth changes controlling struggles to accurately express occlusion, resulting in the disappearance of the purple light cluster and the deformation and merging of the cars. The third example tests the control of forward and backward movements. Compared to our~\method, single-point control is not very sensitive to size changes caused by forward and backward movement. Quantitative results in~\cref{tab:supp1} also show the advantage of 3D motion representation with clustering and dispersion of multiple points. Ablation study in Tab. 2 of the main text indicates that the value of depth does not significantly affect the quality of the generated results. And results in this section show that 2D trajectories with depth value changes can not represent 3D motions. These conclusions both suggest that in our method, the clustering and dispersion of multiple control points are the key aspects of 3D motion representation, while depth information is generally used for moving objects in 3D space to obtain rendered object masks.

\section{Bad Case Analysis}\label{sec:bad}
\begin{figure}[h]
    \centering
    \includegraphics[width=\columnwidth]{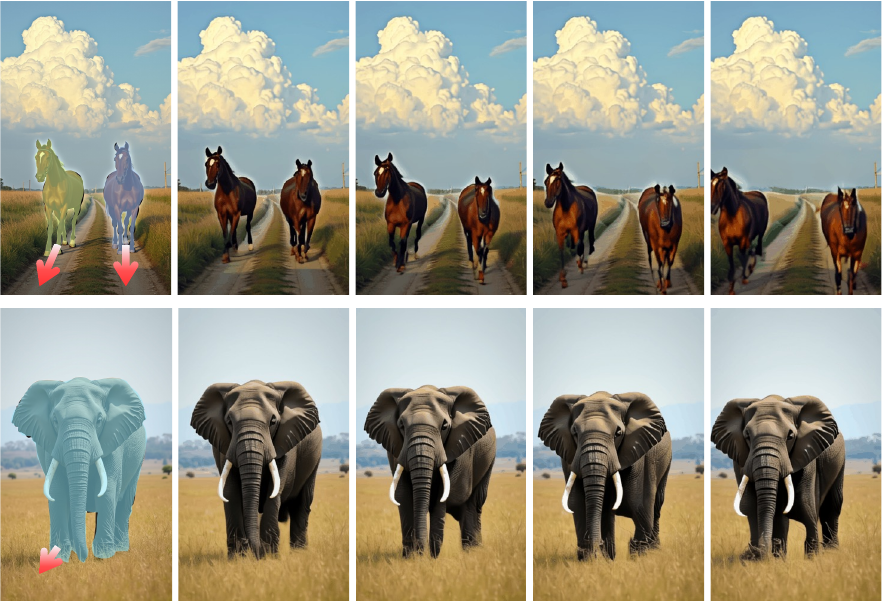}
    \caption{Bad Cases of~\method.}
    \label{fig:supp3}
    \vspace{-5pt}
\end{figure}

In this section, we list some bad generation cases for analysis. Results in~\cref{fig:supp3} indicate that our~\method has difficulties in reconstructing small faces and generating scenes with large motions. It may also confuse similar parts of objects. For example, in the first row of~\cref{fig:supp3}, the horse's faces become blurry while walking, and the movement of their legs is also quite unnatural. In~\cref{fig:supp1}, the movement of the person's feet while running also appears unnatural. In the second row, the elephant's front leg suddenly turns into a back one, and then a regenerated front leg appears. We attribute this phenomenon to the fact that the underlying video base model Stable Video Diffusion (SVD)~\cite{SVD} we apply can not reconstruct small faces and tends to produce artifacts when generating large-scale movements. We are going to enhance our model by integrating more advanced video-based models in the future, hoping to better capture deformable objects and complex dynamics to handle large-scale and non-rigid motions. 

\newpage
{
\small
\bibliographystyle{ieeenat_fullname}
\bibliography{main.bbl}
}

\end{document}